\documentclass{article}

\usepackage[preprint]{neurips_2025}

\usepackage[utf8]{inputenc} 
\usepackage[T1]{fontenc}    
\usepackage{hyperref}       
\usepackage{url}            
\usepackage{booktabs}       
\usepackage{amsfonts}       
\usepackage{nicefrac}       
\usepackage{microtype}      
\usepackage{xcolor}         
\usepackage{graphicx}
\usepackage{subfigure}
\usepackage{booktabs} 
\usepackage{pifont}
\usepackage{soul}
\usepackage{enumitem}
\usepackage{multirow}
\usepackage{makecell}

\usepackage{hyperref}

\usepackage{algorithmic}

\usepackage{algorithm}

\usepackage{amsmath}
\usepackage{amssymb}
\usepackage{mathtools}
\usepackage{amsthm}

\usepackage[capitalize,noabbrev]{cleveref}

\theoremstyle{plain}

\theoremstyle{definition}

\theoremstyle{remark}

\usepackage[textsize=tiny]{todonotes}

\title{Mondrian: Transformer Operators via Domain Decomposition}

\author{%
  Arthur Feeney \quad Kuei-Hsiang Huang \quad Aparna Chandramowlishwaran\\
  University of California Irvine\\
  \texttt{\{afeeney, jackh15, amowli\}@uci.edu} \\
}

\begin{document}

\maketitle

\begin{abstract}
Operator learning enables data-driven modeling of partial differential equations (PDEs) by learning mappings between function spaces. 
However, scaling transformer-based operator models to high-resolution, multiscale domains remains a challenge due to the quadratic cost of attention and its coupling to discretization.
We introduce \textbf{Mondrian}, transformer operators that decompose a domain into non-overlapping subdomains and apply attention over sequences of subdomain-restricted functions.
Leveraging principles from domain decomposition, Mondrian decouples attention from discretization. 
Within each subdomain, it replaces standard layers with expressive neural operators, and attention across subdomains is computed via softmax-based inner products over functions.
The formulation naturally extends to hierarchical windowed and neighborhood attention, supporting both local and global interactions.
Mondrian achieves strong performance on Allen-Cahn and Navier-Stokes PDEs, demonstrating resolution scaling without retraining. 
These results highlight the promise of domain-decomposed attention for scalable and general-purpose neural operators.
\end{abstract}


\newcommand{\domain}{\Omega}
\newcommand{\window}{\Theta}
\newcommand{\integralop}{\mathcal{K}}
\newcommand{\centerdomain}{\Gamma}
\newcommand{\vectorspace}[1]{\mathbb{R}^{#1}}
\newcommand{\funcname}[1]{$\mathrm{#1}$}

\section{Introduction}

Operator learning is transforming the way we solve complex scientific and mathematical problems by introducing data-driven methods to approximate operators between function spaces \cite{kovachki2024operator}. 
These methods have shown significant promise in accelerating the solutions of partial differential equations (PDEs)--the mathematical backbone to modeling physical phenomena in fields such as fluid dynamics, climate science, and materials discovery \cite{pathak2022fourcastnet,azizzadenesheli2024neural}. 
A central challenge in solving PDEs is modeling the multiscale relationships within high-dimensional data, spanning local dynamics to long-range dependencies that evolve across spatial and temporal scales.

Transformers have emerged as a compelling backbone for operator learning due to their ability to capture long-range dependencies via self-attention, motivating recent efforts to adapt transformer architectures to this challenging domain \cite{ho2019axial,Cao2021transformer,hao2023gnot,hao2024dpot,li2024scalable,rahman2024pretrainingcodomainattentionneural, herde2024poseidon, calvello2024continuumattentionneuraloperators}. 
Despite this, direct application of transformers to PDEs remains limited by two main challenges. 
First, attention scales quadratically with the size of the input sequence \cite{vaswani2023attentionneed}. 
For operator learning, where the input naturally comprises the set of points in a function's discretization, the naive application of attention on high-resolution or multiscale domains is computationally intractable. 
Second, standard attention is discretization-dependent, requiring problem-specific designs or retraining across resolutions.

Recent transformer-based neural operators have addressed these limitations by approximations or subquadratic forms of attention to make training feasible \cite{Cao2021transformer, li2024scalable}.
Despite the proliferation of proposed approximations and alternative forms of attention that reduce their complexity, many have yet to see widespread adoption in broader applications due to a common degradation in model accuracy compared to vanilla attention \cite{han2024bridging,choromanski2021rethinking}. 
Furthermore, they may even fail to provide a performance improvement on GPUs when considering realistic problem sizes \cite{dao2022flashattention, dao2023flashattention2}. 
Although there has been significant progress in understanding linear attention and state space models \cite{mamba, mamba2}, state-of-the-art models across vision and language from academia and industry continue to favor exact softmax attention. 
Notable examples include Microsoft's DiNO-V2 \cite{oquab2023dinov2}, Meta AI's segment anything \cite{kirillov2023segany} and Llama \cite{touvron2023llamaopenefficientfoundation} models, and Google's vision transformers scaled to 22 billion parameters \cite{dehghani2023scalingvisiontransformers22}. 
These models, much larger than those that we see in operator learning of PDEs, successfully adopt softmax attention using \emph{hardware efficient} implementations like FlashAttention \cite{dao2022flashattention, dao2023flashattention2}.
We argue that similar strategies can and should be used in operator learning, and that standard softmax attention can be just as powerful in this domain.

This paper introduces \textit{Mondrian}, a framework for transformer-based operator learning that supports:

\begin{enumerate}
    \item \textbf{Self-attention over functions.} Inspired by domain decomposition, Mondrian treats the input as a sequence of subdomain-restricted functions, decoupling attention from discretization. This allows exact softmax attention to be applied over subdomains regardless of resolution.
    \item \textbf{Operator blocks for scientific modeling.} Each subdomain is processed using a neural operator, replacing standard transformer layers with learnable integral kernels. We introduce a mixture integral operator that performs robustly across scales.
    \item \textbf{Local and global attention via multilevel decompositions.} Mondrian extends to hierarchical variants such as windowed and neighborhood attention, enabling scalable modeling of both local and long-range PDE dynamics.
\end{enumerate}

We demonstrate Mondrian on challenging PDE benchmarks including Allen-Cahn and Navier-Stokes. Our results show that Mondrian matches state-of-the-art accuracy, while enabling generalization across resolutions without retraining.
\section{Background} \label{sec:background}

\subsection{Operator Learning}

A commonly explored application of operator learning is approximating the solution of PDEs, though it can be applied more broadly to any data that may be interpreted as a function, such as historical climate data \cite{hersbach2020era5}. 
Consider the Poisson equation as a prototypical example: given a forcing function $f \in L^1$, the goal is to find a solution $u \in H^2$ satisfying $\nabla \cdot \nabla u = f$ inside a bounded domain $\domain \subset \mathbb{R}^n$ with boundary condition $u \equiv 0$ on $\partial \domain$.
The Poisson equation has a solution operator $F : L^1 \rightarrow H^2$, such that $F(f) = u$. 

Operator learning trains a parameterized operator $\hat{F}$ on a dataset of paired forcing functions and their corresponding solutions, such that $\hat{F}(f) \approx u$. 
The learned operator can then be used to approximate solutions for new forcing functions drawn from a similar distribution as the training data.  
Ideally, a learned operator should demonstrate robustness to different discretizations of these functions, though being truly invariant to the discretization cannot be expected \cite{bartolucci2024representation}.

\textbf{Neural Operators (NO)} are a class of neural network designed to approximate operators $O : \mathcal{V}^m \rightarrow \mathcal{U}^n$ between Banach spaces. 
These models are often constructed using approximations of a kernel integral operator $u(x) = (Iv)(x) = \int_\domain \kappa(x, y; \theta) v(y) dy$, where the kernel function $\kappa$ is parameterized by $\theta$ \cite{kovachki2021neural}. 
Numerically evaluating the integral operator is $O(N^2)$ for an $N$ point discretization of $\domain$. 
Since $N$ may be large when applied to an entire problem domain, this has motivated research on reducing the complexity \cite{li2020fourier, li2020neuraloperatorgraphkernel,Cao2021transformer}.

While claims of ``zero-shot super-resolution'' have received fair criticism \cite{kovachki2021neural, bartolucci2024representation, fanaskov2024spectralneuraloperators}, neural operators are able to handle different discretizations seen during training, an important property for field data. 
In practice, they achieve an approximate discretization invariance on real datasets.

\subsection{Domain Decomposition} \label{sec:background-ddm}

Domain decomposition methods are often used as preconditioners for large linear systems arising from the discretization of PDEs.  
These methods typically partition a global domain into subdomains, and iteratively solve problems on those subdomains and recombine their solutions \cite{chan1994ddm, dolean2014ddm}. 
This work does not propose a domain decomposition method, but uses some of the same basic tools. 
For a bounded domain $\domain \subset \mathbb{R}^n$, we use $\mathcal{D}_s(\Omega) = \{ \domain_i \subset \domain : i \in \lbrack s \rbrack\}$ with $\cup_{i \in \lbrack s \rbrack} \domain_i = \domain$  and $\domain_i \simeq \domain_j$ to denote a decomposition of $\domain$ into $s$ subdomains. 
We consider non-overlapping decompositions in which $\text{int}(\domain_i) \cap \text{int}(\domain_j) = \emptyset$ for all $i,j \in \lbrack s \rbrack$ with $i \not= j$. 

Domain decomposition methods are often developed using restriction and extension operators \cite{dolean2014ddm}.
For a function $f : \domain \rightarrow \mathbb{R}^n$ and a decomposition $\mathcal{D}_s(\domain)$, we use subscript $f_{\domain_i} : \domain_i \rightarrow \mathbb{R}^n$ to denote the \emph{restriction} of $f$ to the subdomain $\domain_i$. 
An \emph{extension} operator $E^{\domain_i} : \{\domain_i \rightarrow \mathbb{R}^n\} \rightarrow \{\domain \rightarrow \mathbb{R}^n\}$ performs a zero extension from $\domain_i$ to $\domain$, such that $E^{\domain_i}(f_{\domain_i}) = f(x)$ for $x \in \domain_i$ and zero elsewhere.  
Decompositions of complex domains can be computed using tools like METIS \cite{karypis1998fast}.

\subsection{Transformers}

The transformer was originally introduced for natural language processing \cite{vaswani2023attentionneed}, and subsequently extended to computer vision \cite{dosovitskiy2021imageworth16x16words,liu2021swintransformerhierarchicalvision, liu2022swintransformerv2scaling}. 
With position encoding, transformers are universal approximators for sequence-to-sequence functions \cite{yun2020transformeruniversal}.

One of the core components of a transformer is \emph{attention}, which operates on sequences of data. 
This means it is permutation equivariant and suitable for processing variable-sized inputs. 
Self-attention takes three matrices $Q, K, V \in \mathbb{R}^{l \times d}$ as queries, keys, and values, where $l$ is the length of the sequence and $d$ is the the embedding dimension. 
In computational and I/O complexity analysis of attention, sequence length $l$ is typically the most important variable since it is determined by the input, while $d$ is fixed by the model. 
Scaled dot-product attention takes the form $\text{softmax}(\tau QK^T)V \in \mathbb{R}^{l \times d}$, where the softmax is applied row-wise  \cite{vaswani2023attentionneed}. 
The scalar $\tau \in \mathbb{R}$ is a hyperparameter. 
The matrix $S = \tau QK^T \in \mathbb{R}^{l \times l}$ contains attention scores, with matrix $P = \text{softmax}(S)$ interpreted as probabilities. 


\subsection{Notation}

In the remainder of the paper, we adopt a compressed notation for function spaces: $\mathcal{V}^m = V(\domain; \mathbb{R}^m)$. 
Function spaces are in calligraphic font with the superscript indicating the codomain's dimension. 
The domain ($\domain$ or a subdomain) is assumed to be clear from the context. 
Following the convention in \cite{kovachki2021neural}, $\mathcal{V}^m$ and $\mathcal{U}^m$ denote input and output spaces respectively. 
Additionally, we use $\mathcal{H}^m$ to denote intermediate ``hidden'' spaces (not Hilbert spaces). These symbols are reused across operators and algorithms. A comprehensive notation table is included in Appendix \ref{app:notation}.
\section{Methods} \label{sec:method-model}

\subsection{Sequences of Subdomains}
\label{subsec:seqsubdom}

Transformers used in computer vision typically rely on patching to reduce the size of the input sequence \cite{dosovitskiy2021imageworth16x16words, liu2021swintransformerhierarchicalvision}. 
These methods are resolution-dependent, requiring retraining or finetuning for different input resolutions. 
This makes them unsuitable for operator learning, where the input is the discretization of a function and could be provided at different refinement levels.

\begin{figure}[htbp]
    \centering    \includegraphics[width=0.9\linewidth]{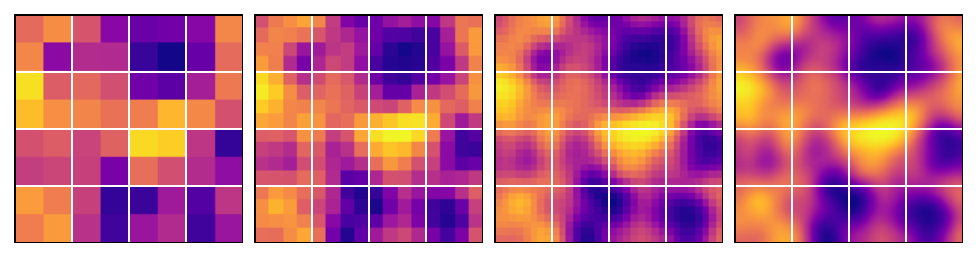}
    \caption{An example domain partitioned into a $4\times 4$ grid of subdomains, delineated by white lines. The spatial extent of each subdomain is determined by the physical size of the domain and is independent of the function's discretization.}
    \label{fig:subdomains}
\end{figure}

Our insight is that the domain of the kernel integral operator can be decomposed to construct function space analogues of transformer architectures that use patching. 
Drawing from domain decomposition methods (outlined in Section \ref{sec:background-ddm}), we decompose a domain $\domain$ into non-overlapping subdomains $\{\domain_i, \dots, \domain_s\}$ and create the transformer's input set using function restrictions $\{v_{\domain_1}, \dots, v_{\domain_s}\}$. 
Each \emph{subdomain} acts as the functional equivalent of a \emph{patch} in vision transformers, but importantly, is independent of the input resolution. 
As shown in Figure \ref{fig:subdomains}, the subdomain sizes only depend on the physical size of the domain.
The model's input is now a sequence of functions, rather than a sequence of vectors. 
Consequently, we must adapt the internals of the transformer, replacing standard operations with neural operators.

\subsection{Design of the Operator Transformer Block}
\label{subsec:modelarch}

The operator form of the transformer block forms our main contribution. 
As many of its components are interchangeable, we discuss each of the following independently:

\begin{enumerate}[topsep=0pt,itemsep=-1ex,partopsep=1ex,parsep=1ex]
    \item Self-attention applied to a sequence of subdomains.
    \item Variants based on two-level decompositions into windows and overlapping neighborhoods.
    \item Multihead attention as an additional level of decomposition.
    \item Choices for ``subdomain operators'' that are used to compute the queries, keys, values, as well as to replace the feed-forward components.
    \item Choices for position encoding.
\end{enumerate}

\textbf{Subdomain Self-Attention.} We describe how to apply self-attention to a sequence of subdomains formed by a non-overlapping decomposition of a domain $\domain$ into $s$ subdomains $\{\domain_1, \dots, \domain_s\}$. 
This differs from standard attention because we must compute attention on sets of functions, rather than sets of vectors. 
Similar approaches for attention on functions have been proposed \cite{rahman2024pretrainingcodomainattentionneural, calvello2024continuumattentionneuraloperators}. 
We use a neural operator $I_{QKV} : \mathcal{V}^d \rightarrow \mathcal{H}^{3d}$ between two function spaces to compute the triplet of functions $(Q_i, K_i, V_i) = I_{QKV}(v_{\domain_i})$ for each subdomain $i$. 
The triplet is formed by splitting along the codomain of the operator's output.
The next step is to compute the score matrix $S \in \mathbb{R}^{s \times s}$. 
The components of the score matrix are computed with a function inner product of the queries and keys: $S_{k,j} = \langle Q_k, K_j \rangle$. 
We use the $L_2$ inner product. 
Note that $S$ is a finite matrix, not a function. 
The function-space attention is then defined for each subdomain, similar to standard attention as: $u_{\domain_i} = \text{softmax}(\tau S)_iV$, where $\tau$ is a temperature parameter and the softmax is applied row-wise. 
Finally, the results from each subdomain are recombined to produce the output function $u$ over the entire domain.
The full procedure is summarized in Algorithm \ref{alg:function-attention}.

\begin{algorithm}
\caption{Subdomain-Self-Attention}\label{alg:function-attention}
\hspace*{0.5em} \textbf{Input}  $v : \domain \rightarrow \mathbb{R}^m$, number of subdomains $s$ \\
\hspace*{0.5em} \textbf{Output} $u : \domain \rightarrow \mathbb{R}^m$
\begin{algorithmic}[1]
\STATE $\{\domain_1, \dots, \domain_s\} = \mathcal{D}_s(\domain)$
\STATE $(Q_i, K_i, V_i) = I_{QKV}(v_{\domain_i})$, for $i \in \lbrack s \rbrack$
\STATE Compute S with $S_{k,j} = \langle Q_k, K_j \rangle$, for $k, j \in \lbrack s \rbrack$
\STATE $u_{\domain_i} = \text{softmax}(\tau S)_iV$, for $i \in \lbrack s \rbrack$
\STATE $u = \sum_{i \in \lbrack s \rbrack} E^{\domain_i}(u_{\domain_i})$ (Recompose $\domain$)
\end{algorithmic}
\end{algorithm}

\textbf{Neighborhood and Windowed Attention.} Neighborhood and windowed attention can be viewed as a two-level hierarchical decomposition of the domain $\domain$. 
Windowed attention is summarized in Algorithm \ref{alg:window-function-attention}.
We first decompose the domain into $w$ larger \emph{windows} $\{\Theta_1, \dots, \Theta_w\}$.
Within each window, we perform the \funcname{Subdomain\text{-}Self\text{-}Attention} with a second level of decomposition in each window.
Each window is processed in parallel, with the results from all windows recomposed to form the output function $u$ over the entire domain.

\begin{algorithm}
\caption{Windowed-Func-Attention} \label{alg:window-function-attention}
 \hspace*{0.5em} \textbf{Input}  $v : \domain \rightarrow \mathbb{R}^m$, number of windows $w$\\
 \hspace*{0.5em} \textbf{Output} $u : \domain \rightarrow \mathbb{R}^m$
\begin{algorithmic}[1]
\STATE $\{\Theta_1, \dots, \Theta_w\} = \mathcal{D}_w(\domain)$ (Decompose into windows)
\FOR{window $i \in \lbrack w \rbrack$}
\STATE $u_{\Theta_i}$ = Subdomain-Self-Attention($v_{\Theta_i}$)
\ENDFOR
\STATE $u = \sum_{i \in \lbrack w \rbrack} E^{\Theta_i}(u_{\Theta_i})$ (Recompose $\domain$)
\end{algorithmic}
\end{algorithm}

To allow information to propagate across window, we can apply shifting similar to Swin transformers \cite{liu2021swintransformerhierarchicalvision}. 
Similar to how the windows and subdomains are created, the amount of shifting depends on the physical size of the domain and not its resolution. 
For each attention block, the domain is cyclically shifted by a fixed amount to get the shifted configuration. 
Windowed function attention is then applied on the shifted configuration of $\domain$, using an updated attention mask to account for the new window arrangement. 

Neighborhood attention is conceptually similar but differs in the use of \emph{overlapping} decompositions, allowing information to propagate. We defer to Appendix \ref{app:models} for its detailed description.

\textbf{Multihead Attention.} 
In standard transformers, multihead attention is typically implemented by partitioning along the embedding dimension. 
For example, if the query is a tensor $q \in \mathbb{R}^{b \times s \times e}$ (where $b$ is the batch size, $s$ is the sequence length, and $e$ is the embedding dimension), multihead attention splits $q$ into $h$ heads along the embedding dimension: $h \in \mathbb{R}^{b \times h \times s \times (e / h)}$.

Since our input is a sequence of functions defined on subdomains, multihead attention can take different forms. 
One way is to partition along the codomain \cite{calvello2024continuumattentionneuraloperators}. 
Alternatively, we can partition the domain spatially by performing another level of decomposition. Thus, we may have two or three levels of decomposition. 
While partitioning along the codomain is well-suited for high-dimensional problems, spatial decomposition might be advantageous for spatially heterogeneous problems where local interactions across different regions of the domain are important.
A hybrid approach could combine the strengths of both methods.

\textbf{Subdomain Operators.} There are two instances where we apply neural operators on subdomains. 
The first is the operator $I_{QKV}$, described earlier, which computes the queries, keys, and values.
The second instance is in the \emph{feed-forward} layers, where we define the operation as a pair, $I_1$ and $I_2$, of operators: $I_2\lbrack\sigma(I_1v + b_1)\rbrack + b_2$. This is clearly analogous to the typical feed-forward layer using linear weights \cite{vaswani2023attentionneed}. The function $\sigma$ is an activation and $b_1$, $b_2$ are bias functions. Similar to the bias of a convolutional layer, which learns on vector for every point of the output, the biases $b_1$ and $b_2$ are constant functions of space. 

The choice of subdomain operators is important for the performance and quality of the model. Global linear operators, such as Fourier Neural Operator (FNO)'s spectral convolution are natural candidates due to their similarity to the finite linear layers in transformers. Much of the research on neural operators has explored strategies to achieve subquadratic complexity \cite{kovachki2021neural, li2020fourier, li2020neuraloperatorgraphkernel}. In this work, since we assume that we are only working with small subdomains, it is reasonable to use subdomain operators with quadratic complexity. One example of this would be to directly implement a parameterized integral operator \cite{kovachki2021neural}.

We experiment with multiple choices of subdomain operators:

\begin{itemize}
    \item A standard neural operator that is implemented with an integral operator, which has the form $\int_D \kappa(x, y)v(y) dy$
    \item Low Rank Integral Operator, which has the form $\int_D \kappa_1(x)\kappa_2(y)^T v(y)dy$. The kernel functions $\kappa_1$ and $\kappa_2$ return matrices in $\mathbb{R}^{m\times r}$, so the rank of the kernel is at most $r$.
    \item Spectral convolutions \cite{li2020fourier} are efficient for large domains; however, we encountered problems when trying to apply them to smaller subdomains.
    \item Mixture Operator, which defines $\kappa(x, y)$ as $\sum_{i \in \lbrack l \rbrack} M_iC_i(x, y)$. The $M_i \in \mathbb{R}^{m\times n}$ are learnable matrices and $C : \domain \times \domain \rightarrow \mathbb{R}^l$ is a small neural network that produces scales. The hyperparameter $l$ determines the number of matrices that are mixed.
\end{itemize}

We include ablation experiments for the different subdomain operators in Appendix \ref{app:ablations}.

\textbf{Position Encoding.} Similar to general transformers, the choice of position encoding is somewhat flexible. 
Since we assume the input is a sequence of subdomains instead of patches, in some cases, the position encoding needs to be an operator.

Similar to the original vision transformer, we can use a learned position encoding for each subdomain. 
In our case, the encoding applied to each subdomain $i$ must also be an operator $P_i : \mathcal{V}^m \rightarrow \mathcal{U}^m$, which adds a position encoding to the input function: $(P_iv_{\domain_i})(x) = v_{\domain_i}(x) + p_i(x; \theta)$, where $p_i : \domain_i \times \Theta \rightarrow \mathbb{R}^m$ is parameterized by $\theta$. 
We found that enforcing $p_i$ to be a spatial constant works well, simplifying the encoding to $(P_iv_{\domain_i}) = v_{\domain_i}(x) + p_i$, where $p_i \in \mathbb{R}^m$ is a learned vector.

Another approach to position encoding is the relative position bias \cite{shaw2018self}, which is unchanged from typical transformers. 
Since the attention scores still form a finite matrix, we can compute the entries as $S_{k, j} = \langle Q_k, K_j\rangle + B_{k,j}$, for a learned matrix $B$ that encodes relative distances between subdomains. In practice, $B$ can be constructed in a similar fashion to the Swin transformer's position bias \cite{liu2021swintransformerhierarchicalvision}. In practice, we found this worked suitably well for our problems and is what is used in our experiments.

Finally, rotary position encodings \cite{su2024roformer} are gaining popularity and could also be implemented simply by applying the rotation along the function's codomain.
Though we do not explore this approach in this work, we believe it is interesting for future research.

\subsection{Extensions and Variations} \label{sec:extensions}

\textbf{Localized Cross Attention.} One advantage of working with subdomains is the ability to efficiently map between different discretizations by using a neural operator or cross attention on the points within each subdomain. 
In this approach, the query $Q$ is constructed from the desired discretization. 
The key and value $K,V$ are constructed from data on the existing discretization. 
By computing attention, we can directly interpolate onto a new grid. 
This method is useful for converting an irregular discretization to a regular grid, and can also serve as an expressive coarsening or refinement layer. 
Since this is applied to each point in the discretization, it becomes computationally expensive for the full domain but works well in subdomains. For future work, this concept could potentially be useful for  adaptive refinement \cite{berger1984adaptive,berger1989local}, where some subdomains may benefit from using a finer discretization. This idea is similar to the geometry informed neural operator \cite{li2023geometryinformed}, which uses a graph neural network to project a complex domain to a regular grid, so that it can be processed with a FNO.

\textbf{UNet-Style Architecture.} A UNet-style architecture can be constructed with a linear complexity in the domain size and the number of subdomains. 
This would be similar to the hierarchical Swin transformer \cite{liu2021swintransformerhierarchicalvision}. 
However, the method of downsampling may differ. 
Downsampling to a coarser grid could be done using cross-attention, a subdomain operator, or simply by smoothing the hidden representation and then directly downsampling \cite{raonic2024convolutional}.

\label{subsec:flash}


\section{Experiments} 
\label{sec:experiment}

\subsection{Problems and Datasets}
We evaluate Mondrian on time-dependent PDEs by learning their solution operators as surrogates for numerical solvers. 
Such operator surrogates for dynamical systems have great potential due to their ability to take relatively larger timesteps compared to traditional solvers, reducing modeling time with reasonable accuracy. 
We focus on 2D problems because, generally, time-dependent 3D simulations generate huge outputs that are not stored long-term and only the derived quantities of interest are stored (instead of full-field data). The Allen-Cahn dataset also contains data generated at different resolutions. This allows us to study resolution scaling and is useful for ablation studies to isolate the impact of different design choices in our transformer models. We compare many models and perform extensive hyperparameter tuning on the Allen-Cahn dataset, with results in Appendix \ref{app:allen-cahn-exp}. Additionally, we consider a challenging Navier-Stokes dataset in a 2D domain with decaying turbulence simulated using a spectral solver in JAX-CFD \cite{dresdner2022spectral-jaxcfd}. 

All models are trained on a Nvidia A30 GPU. Training a single model on the Allen-Cahn dataset takes about two hours. The overall time for hyperparameter tuning on the Allen-Cahn dataset was several hundred GPU hours, since we considered multiple configurations for each baseline. Training on the Navier-Stokes dataset takes about twelve hours per model. 

\subsection{Baseline Models}
Many of the recently proposed transformer architectures for operator learning are quite specific to different types of problems. 
For instance, rather than proposing a new style of attention, some models focus on the ability to generalize across different categories of simulation \cite{hao2024dpot,herde2024poseidon}, process different sequences of variables \cite{rahman2024pretrainingcodomainattentionneural}, or generalize to various geometries \cite{hao2023gnot,hao2024dpot}. 
This makes it difficult to directly compare forms of attention in other transformer operators. Furthermore, many of these choices are orthogonal to the backbone architecture. Since our main contributions are local and global attention operators, we focus our experiments on evaluating the impact of these layers.

The following baseline models are chosen for comparison:
\begin{itemize}[topsep=0pt,itemsep=-1ex,partopsep=1ex,parsep=1ex]
    \item Fourier Neural Operator (FNO) \cite{li2020fourier}.
    \item Factorized Fourier Neural Operator (FFNO) \cite{tran2023factorized}.
    \item Galerkin Transformer (GT) uses a softmax-free form of attention with linear complexity \cite{Cao2021transformer}.
    \item FactFormer (FF) does a similar strategy to FFNO, and computes attention along each axis of the domain \cite{li2024scalable}. This is similar to an Axial transformer \cite{ho2019axial}.
    \item Fourier Attention Neural Operator (FANO) \cite{calvello2024continuumattentionneuraloperators}. 
\end{itemize}

Details on the architectures and hyperparameters of these models can be found in Appendix \ref{app:models}.

\subsection{Allen-Cahn: Phase Field}

\begin{figure}[h]
    \centering
    \includegraphics[width=1\linewidth]{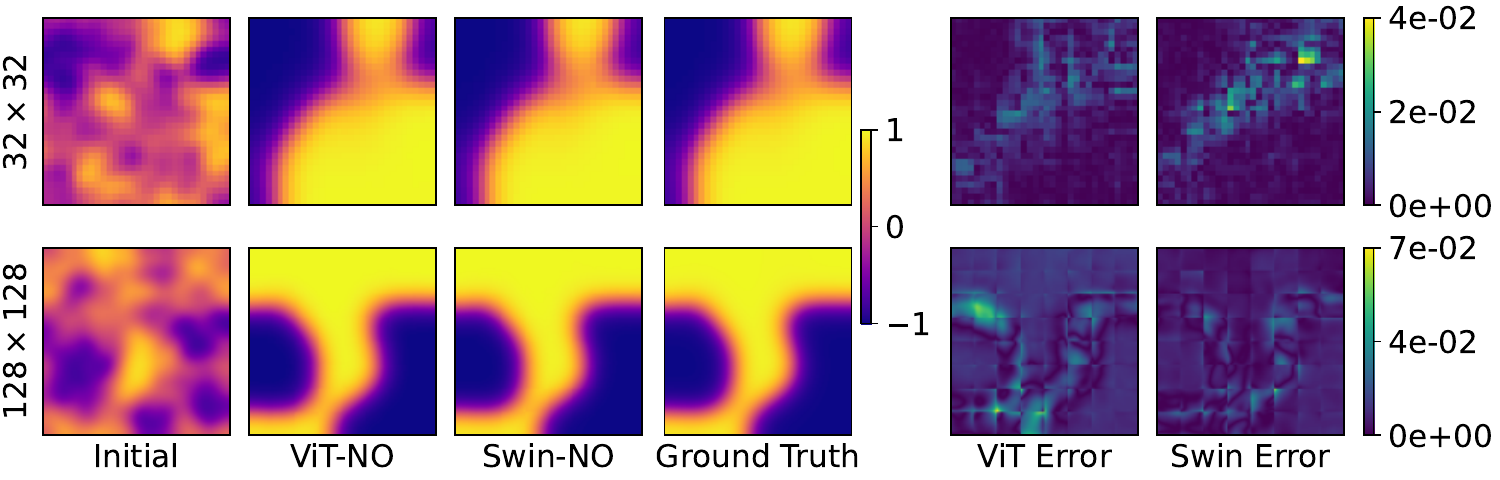}
    \vspace{-2em}
    \caption{Mondrian operator predictions on Allen-Cahn using the Mixture Operator as the subdomain operator, with the pointwise absolute error between the model and ground truth. Top row: sample output from the test set, with the same 32$\times$32 resolution as the training set. Bottom row: A sample from the test set at 128$\times$128 resolution demonstrates the models ability to interpolate higher resolutions without requiring retraining.}
    \label{fig:ac-visual-result}
\end{figure}

 
The Allen-Cahn equation is a semi-linear PDE, originally proposed to model phase separation in binary alloys \cite{allen1976mechanisms}, is given by: $u_t = \gamma \Delta u - f(u)$. We use the common double-well potential $f(u) = u^3 - u$ and enforce Neumann boundary conditions ($\partial_n u = 0$) on $\partial \domain$. The parameter $\gamma$ controls the interfacial width. The initial condition is a Gaussian random field with decaying power spectrum. We restrict the field to the range $\lbrack -0.5, 0.5 \rbrack$, so each point can be interpreted as a mixture of two phases. Over time, the system approaches metastable states of $+1$ and $-1$. The model takes the initial condition and $\gamma \in \lbrack 1\times 10^{-4}, 5\times 10^{-3} \rbrack$ as input and estimates the solution at time $t = 6$. An example is shown in Figure \ref{fig:ac-visual-result}. 


We construct a training dataset of 20,000 simulations at $32 \times 32$ resolution and use a 70/30 train-validation split. The test set consists of 6000 held-out simulations, evenly split across resolutions of $32 \times 32$, $64 \times 64$, and $128 \times 128$. This setup enables us to rigorously evaluate resolution generalization and interpolation.

As shown in Figure~\ref{fig:ac-visual-result}, Mondrian successfully captures the metastable interface dynamics at both training and higher resolutions. 
Models using the Mixture Operator maintain low error even at $128 \times 128$ resolution, despite only being trained on $32 \times 32$ data.
This demonstrates Mondrian's ability to decouple modeling capacity from discretization scale. Additional experiments and more complete results are in Appendix \ref{app:allen-cahn-exp}.

\subsection{Navier-Stokes: Turbulent Flow}

Modeling turbulent flow presents a significant challenge for neural operators. We evaluate Mondrian on this problem using data generated from JAX-CFD spectral solver \cite{dresdner2022spectral-jaxcfd}. We simulate a 2D periodic domain $\lbrack 0, 2\pi \rbrack^2$ with viscosity $0.001$ and maximum velocity $3$. 
To generate the training data, we run simulations for 10s using a timestep of 0.004s, yielding 2500 simulation steps in total. We save 50 frames per simulation, corresponding to one frame every 50 simulation steps (i.e., every 0.2s).
The data is normalized such that the standard deviation of values is approximately one.
For training, the model receives two input frames and predicts two frames into the future. 

\begin{figure}
    \centering
    \includegraphics[width=0.8\linewidth]{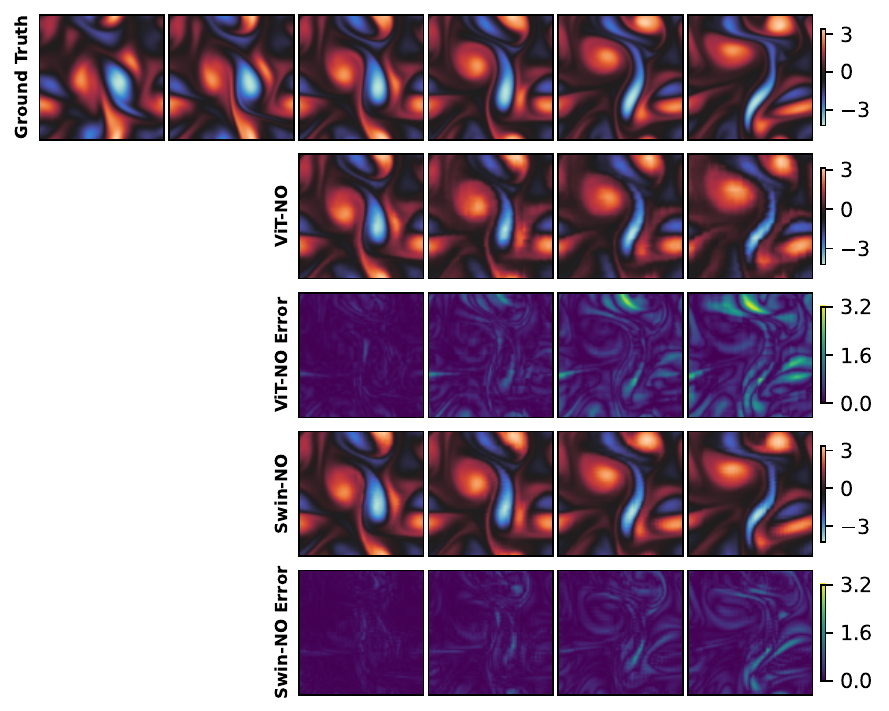}
    \caption{Vorticity field prediction rollouts of turbulent flow across different timesteps ($\Delta t$ = 0.2s) using the Mondrian operators with the best performing subdomain operator. We show the pointwise absolute error between the ground truth and prediction at each timestep.}
    \label{fig:ns-vit-spectral}
\end{figure}

There are several challenges with modeling the turbulent Navier-Stokes equations. (1) The dynamics are highly sensitive, so two very similar inputs can have noticeably different solutions. So, the training data is essentially always a noisy form of the true solution. (2) The solution contains fine-scale structures (e.g., high-frequency vortices), which many neural models struggle to represent accurately. 

Attention, including those used in Mondrian, have been observed to act as a low-pass filter \cite{wang2022antioversmoothing}, which may suppress high frequency components important to modeling turbulence.
As shown in Figure \ref{fig:ns-vit-spectral}, the predicted vorticity fields appear smoothed, with sharp vortices visibly reduced compared to the ground truth. This indicates that although Mondrian captures the coarse-scale structure well, it struggles to preserve high-frequency details over multiple rollout steps.
Addressing the spectral bias introduced by attention layers remains an important direction for future work.
\section{Related Work}
Two prior works are closely related to ours. \emph{Continuum attention} \cite{calvello2024continuumattentionneuraloperators} is recent work that develops a ViT-style operator that, like our approach, uses extension and restriction operators. We go in somewhat different directions: they make large contributions to the analysis of these models, while our focus is more on practical design. \emph{How to Understand Masked Autoencoders} \cite{cao2022understandmaskedautoencoders} studies the effectiveness of masked image pretraining \cite{he2022masked} using an integral kernel under a non-overlapping domain decomposition. However, it does not explore its application as a neural operator.

Our contributions lie in the design of an operator transformer block with multi-level decompositions and efficient local attention operators. Transformers have garnered much interest because of their ability to scale, but without an efficient attention, this scaling is not possible.

\textbf{Connection To Domain Decomposition Methods.} While this work is not intended to be used with or in place of numerical domain decomposition methods \cite{dolean2014ddm}, some analogies can be drawn. One is between the neighborhood attention and the Alternating Schwarz Method (ASM) \cite{lions1988schwarz}. ASM solves subdomains iteratively and relies on the subdomains' overlap to propagate information across the domain. The neighborhood attention uses a similar overlapping strategy to enable information propagation. Both require $n$ steps to propagate information across $n$ subdomains.  Furthermore, The convergence of ASM can be improved using a coarse, global solve. This enables global information to propagate more quickly \cite{dolean2014ddm}. This is actually similar to UNet-Style architectures that perform pooling and upscaling to ensure the model is applied globally to the input.

There are also non-overlapping domain decomposition methods, such as Optimized Schwarz Methods \cite{gander2006optimized,dubois2012optimized}. While often approximated locally, in theory, optimal transfer rates depend on the global information. There is some similarity with the ViT operator, in which the transformation applied to a subdomain depends globally on the other subdomains.



\textbf{Neural Operators.} There are a large number of neural operators that propose subquadratic approximations of the integral operator \cite{li2020fourier,kovachki2021neural,Cao2021transformer}. This work takes a slightly different approach to enable use of computationally expensive, but expressive and proven attention methods in operator learning \cite{vaswani2023attentionneed,dosovitskiy2021imageworth16x16words,liu2022swintransformerv2scaling,liu2021swintransformerhierarchicalvision}.

There are neural operators that break the domain into pieces to create efficient models. The Graph Kernel Neural operator \cite{li2020neuraloperatorgraphkernel} applies a kernel integral over balls around each point: $u(x) = \int_{B(x)}\kappa(x, y)v(y)dy$, where $B(x)$ is a ball around $x$ and $x \in \domain$. The neighborhood attention is similar to this, but we consider a radius of subdomains around subdomains. Recent work on localized \cite{schiaffini2024localizedintegral} uses DISCO convolutions to construct local versions of integral operators. Convolutional neural operators use a sinc filter to up and down sample. DPOT uses a similar strategy to downsample the input to a transformer \cite{hao2024dpot} This form of interpolation has the potential downside of being somewhat problem specific; not all fields make sense to interpolate with a sinc filter. 

There is existing work that proposes performing attention on subdomains. \cite{rahman2024pretrainingcodomainattentionneural} computes attention across the function codomain, enabling handling multiple problems with different variables. As mentioned, \cite{calvello2024continuumattentionneuraloperators} takes a similar approach using function restrictions and extensions, but we arrive at a different construction.




\section{Conclusion}
In this work, we present \textbf{Mondrian}, efficient operator blocks of popular transformers such as ViT and Swin designed to be independent of the input resolution.

\textbf{Limitations.} First, our work inherits limitations of current neural operator paradigms \cite{bartolucci2024representation}. Zero-shot superresolution (i.e., recovering missing features without finetuning) remains challenging. 
Any dataset of PDE solutions will be drawn from a distribution that can be approximated by a finite dimensional space. 
However, the full set of solutions to a PDE will not have a finite structure.
As a result, neural operators often fail to reconstruct fine-scale features when applied to high-resolution inputs, especially when those features lie outside the training distribution.
This contrasts with image models, where low-dimensional manifold assumptions often justify upscaling \cite{pope2021intrinsic}.

Second, we rely on training the model to combine learned subdomain dynamics. As the physical domain size increases, new subdomain behaviors may emerge that are not represented in training, limiting scalability to arbitrarily large domains without additional data or finetuning.

Third, FlashAttention is restricted to embedding dimensions that are fairly small powers of two, so operations map well to GPU hardware \cite{dao2022flashattention}. For general machine learning, this is not an issue, as one can always round up the embedding dimension to a power of two. This is non-trivial in our setting, where the number of points in a subdomain can vary. However, as discussed in Section \ref{sec:extensions}, we can mitigate this by projecting each subdomain to a regular grid with a fixed number of points. 

Finally, designing replacements for the standard linear layers in transformers remains an open question. Our Mixture Operator and other subdomain integral kernels offer promising alternatives, but further exploration of learned basis functions and structured kernels is warranted.

\textbf{Broader Impact.}  We are not aware of directly harmful applications of the problems discussed in this paper. However, PDEs have been the most powerful tool to model physical systems for decades. Their use is very widespread and has potential for both societal good and harm. For instance, PDEs are heavily used in the energy industry to model and design wind turbines. Similarly, PDEs are also used by the oil and gas industry to model how liquids flow through the porous media found in subterranean structures.

\textbf{Looking Forward.} There are many opportunities to extend this work. Mondrian provides a modular blueprint for constructing transformer-based neural operators that are resolution-agnostic and hardware-scalable. They can be easily adapted to derive new transformer variants. 
Other immediate directions include extending our approach to handle different domain sizes, as well as irregular grids and point clouds.

\begin{ack}
This work is supported by the National Science Foundation under the award number 2211908. We thank the Research Cyberinfrastructure Center at the University of California, Irvine for the GPU computing resources on the HPC3 cluster. We thank Ying Wai Li and Yen Ting Lin for helpful discussions on operator learning.
\end{ack}

\bibliography{main}
\bibliographystyle{abbrv}



\newpage
\appendix

\section{Notation and Terminology} \label{app:notation}

\begin{table}[h]
    \centering
    \begin{tabular}{c|p{10cm}}
        $\domain \subset \mathbb{R}^n$ & A bounded domain \\
        $\mathcal{D}_s(\domain)$ & A decomposition of the domain $\domain$ into $s$ non-overlapping subdomains \\
        $\domain_i$ & The $i$-th subdomain of a decomposition \\
        $\Theta_i$ & The $i$-th window used in window attention \\
        $f_{\domain_i}$ & The restriction of a function $f : \domain \rightarrow \mathbb{R}^n$ to the subdomain $\domain_i$. \\
        $E^{\domain_i}$ & The extension operator that extends a function's domain with zeros. \\
        $\mathcal{V}^m$ & This is a compressed notation for function spaces $V(\domain; \mathbb{R}^m)$, since we only use real numbers and the domain is typically obvious from context. We generally use $\mathcal{V}^m$ for input spaces, 
        $\mathcal{U}^m$ for output spaces, and $\mathcal{H}^m$ for ``hidden'' spaces \\
        $I$ & An kernel integral operator: $(Iv)(x) = \int_\domain \kappa(x, y)v(y)dy$. These are distinguished by a subscript. For instance, $I_{QKV}$ is the operator used to compute the queries, keys, and values. \\
        $\lbrack n \rbrack$ & The set of integers $\{1, \dots, n\}$. \\
    \end{tabular}
    \caption{Notation.}
    \label{tab:notation}
\end{table}
\section{Model Details} 

\subsection{Subdomain Operators} \label{subsec:model:subdomop}

In our transformer architectures, we refer to the operators that compose the feed-forward layers and compute the queries, key, and values as ``subdomain operators.'' These subdomain operators essentially replace the linear operations. All of the subdomain operators are based on integral operators $(Iv)(x) = \int_{\Gamma} \kappa(x, y; \theta)v(y)dy$. We found the choice of subdomain operator to be extremely important.

\begin{itemize}[topsep=0pt,itemsep=-1ex,partopsep=1ex,parsep=1ex]
    \item \textbf{Vanilla Integral Operator}. In this case, $\kappa(x, y): \Gamma \times \Gamma \rightarrow \mathbb{R}^{m\times n}$, where $\kappa$ is a small neural network that maps the input coordinates to a matrix. Evaluating this operator for an $N$ point discretization of $\Gamma$ requires $O(N^2)$ computations.

    \item \textbf{Low-Rank Integral Operator.} The low-rank integral operator assumes $\kappa(x, y) = \phi(x)\psi(y)$, where $\phi: \Gamma \rightarrow \mathbb{R}$ \cite{kovachki2021neural}. This allows $\phi$ to be factored out of the integral: $Iv = \phi \int_\domain \psi(y)v(y)dy$.
    
    \item \textbf{Spectral Convolution.} The spectral convolution is used in FNO and is parameterized in Fourier space. This is $O(N \log N)$ for an $N$-point discretization (assuming the number of modes used is much smaller than the discretization). The spectral convolution uses the Fourier transform to map the input data to Fourier space, truncates the high frequencies above some cutoff, applies the parameters, and then uses an inverse Fourier transform to map back to real space. Since the higher frequencies are truncated, this operation is a projection to a finite dimensional space. This is essentially downsampling, applying the weights, and then interpolating back to the original resolution. It is necessary to include a skip connection to retain information.

    \item \textbf{Interpolating Integral Operator.} A downside of the integral operators that use a neural network $\kappa$ as a function of the input coordinates is that it forces the model to learn an additional position encoding. Based on the understanding that most neural operators project to a finite space \cite{li2020fourier, bartolucci2024representation}, we propose a simpler operator, \emph{interpolating neural operator} that avoids learning a separate position encoding. This is similar to recent work on integral neural networks \cite{solodskikh2023integralnetworks}, where we essentially interpolate a weight matrix to get an integrable function. The main novelty lies in our construction, which has two parts: (a) We use a low-rank kernel function, in which $\kappa(x,y) = \phi(x)\psi(y)$ \cite{kovachki2021neural}. (b) The functions $\phi$ and $\psi$ are determined as the interpolation of a finite point set. Similar to the low rank operator, this becomes a sum of basis functions $u = \sum_{i\in \lbrack r \rbrack} \phi_r \langle \psi, v\rangle$, where the basis functions $\phi_r$ are determined by the interpolation method. This is similar to the form that implements $\kappa$ as a neural network, but has the advantage of not needing to learn an additional position encoding. We just have a finite set of parameters that are interpolated. As shown in Table \ref{tab:ac-in-dist-test-metrics}, we found this worked very well at the training resolution, but performed poorly when trying to apply on higher resolutions.

    \item \textbf{Attention Operator.} We experiment with using attention inside subdomains (in addition to using attention between subdomains.) This can be implemented by treating the discretization as the input sequence, computing the queries, key, and values on every point and then using masked attention corresponding to the subdomains. 

    \item \textbf{Mixture Operator (MO).} We propose an alternate operator where the kernel function $\kappa$ is constructed as a weighted sum of $n$ trainable matrices $M_1, \dots, M_n$, with the weights learned by a very small neural network $C : \domain \times \domain \rightarrow \mathbb{R}^n$ that outputs $n$ coefficients. Thus, we define $\kappa(x,y) = \sum_{i = 1}^n M_iC_i(x,y)$. We call this a mixture operator because it ``mixes together'' several parameter matrices. We found that operators using this kernel perform extremely well at the training resolution and are stable when evaluating at higher resolutions. The choice of $n$ is a hyperparameter. Results in Table $3$ highlight that this method is robust across different resolutions.  We implement this assuming ``full rank'' and thus the required compute time is $O(N^2)$ for an $N$-point discretization. Since this is only applied inside subdomains, where the discretization will be small, this cost is not unreasonable. We also experiment with a separable version of this architecture, in which the matrices $M_i$ are all diagonal. After computing the integral operator, we apply a linear operation point-wise to mix across channels. This significantly reduces the number of parameters.

    \item \textbf{Separable Mixture Operator.} We consider a separable version of the mixture operator, similar to separable convolutions \cite{chollet2017xception, li2020fourier}. This is used to reduce the number of parameters. In this case, the kernel function $\kappa(x, y) = \sum_{i=1}^n D_iC_i(x, y)$ is a sum of diagonal matrices $D_i$. This will ``mix'' data spatially across the domain, but it will not across separate channels. To mix across channels, we then use a point-wise convolution operation following. 
\end{itemize}

\subsection{Neighborhood Attention}

Neighborhood attention can be expressed in a few ways. One way to implement it would be as a two-level overlapping domain decomposition. A simpler way to express it is as a one-level decomposition, but with attention scores zerod based on the distance between subdomains. Since the subdomains are determined by physical dimensions, setting a number of fixed neighbors to compute attention between will essentially correspond to the neighborhood radius. This looks similar to the standard attention.

\begin{algorithm}
\caption{Neighborhood-Subdomain-Self-Attention}\label{alg:neighborhood-function-attention}
\hspace*{0.5em} \textbf{Input}  $v : \domain \rightarrow \mathbb{R}^m$, number of subdomains $s$, number of neighboring subdomains $r$ \\
\hspace*{0.5em} \textbf{Output} $u : \domain \rightarrow \mathbb{R}^m$
\begin{algorithmic}[1]
\STATE $\{\domain_1, \dots, \domain_s\} = \mathcal{D}_s(\domain)$
\STATE $(Q_i, K_i, V_i) = I_{QKV}(v_{\domain_i})$, for $i \in \lbrack s \rbrack$
\STATE Compute S with components $
S_{k, j} = \begin{cases}
 \langle Q_k, K_j \rangle & \text{if } \text{distance}(\Omega_k, \Omega_j) \leq r \\
 0 & \text{ otherwise} 
\end{cases}
$ for $k, j \in \lbrack s \rbrack$

\STATE $u_{\domain_i} = \text{softmax}(\tau S)_iV$, for $i \in \lbrack s \rbrack$
\STATE $u = \sum_{i \in \lbrack s \rbrack} E^{\domain_i}(u_{\domain_i})$ (Recompose $\domain$)
\end{algorithmic}
\end{algorithm}

\subsection{Baseline Models} \label{app:models}

\textbf{Fourier Neural Operator (FNO)} applies the convolution theorem to the kernel integral operator in order to directly parameterize in Fourier space \cite{li2020fourier, kovachki2021neural}. This enables approximating an $N$ point discretization of the integral operator in $O(N \log N)$ time. Notably, the number of parameters in one layer of FNO scales with $O(H^2 M^D)$, where $H$ is the hidden size, $M$ is the number of modes and $D$ is the number of dimensions. In practice, FNO seems to require a relatively large number of parameters \cite{rahman2024pretrainingcodomainattentionneural}.

Since none of our problems are defined on periodic domains, we always use domain padding with $1/4$ of the domain size \cite{kovachki2021neural}. We also follow common advice that the number of modes should be (approximately) $1/2$ to $2/3$ the training resolution. For example, on a $32\times 32$ domain, we tune with between 16 and 24 modes. We always use group normalization.

\textbf{Factorized Fourier Neural Operator (FFNO)} applies one-dimensional fourier transforms along each axis of the input domain and applies weights to these 1D transformations  \cite{tran2023factorized}. When compared with the standard FNO, this significantly reduces the number of parameters to $O(H^2MD)$. We follow the same guidelines as FNO when setting the number of modes.

\textbf{Galerkin Transformer (GT)} \cite{Cao2021transformer} is based on linear attention and attempts to make linear attention suitable for operator learning. The original version includes additional modules such as CNN-based up/down sampling before the input and applying FNO's spectral convolutions on the output. Since we are primarily interested in how well the linear-style attention works in this setting, we evaluate it in isolation. 
We implement a ``pure'' version of a Galerkin transformer,  that consists only of transformer blocks that use the Galerkin-style linear attention. This allows for a more direct comparison with our models and FactFormer, both of which do not need additional modules to work well.

\textbf{FactFormer (FF)} architecture shares similarity with axial attention \cite{ho2019axial} and FFNO. We reuse the official implementation of Factformer with one caveat: The implementation of FactFormer is not setup as a neural operator (there is an explicit parameter for the resolution in the up/down blocks). For the sake of fairness, we report two results for the $32 \times 32$ resolution data: one is a model that uses the official implementation with the up/down blocks and cannot be applied to higher resolutions. The second omits the up/down blocks, but can be used on higher resolutions.

\section{Allen-Cahn Experiments} \label{app:allen-cahn-exp}

\subsection{Data Preparation}

These experiments for Allen-Cahn are closer to the more traditional experiments for neural operators. Our main goal is essentially to demonstrate that our models work well at the training resolution and have a fairly stable error when moving to finer discretizations. 

The training dataset consists of $20,000$ simulations with random initial conditions and interacial widths randomly chosen from the range $\lbrack 1\times 10^{-5}, 5\times 10^{-3} \rbrack$. The initial conditions are generated by sampling from a Gaussian random field, generated with Pylians \cite{Pylians}, and clipping the distribution to the range $\lbrack -0.5, 0.5 \rbrack$. This can be interpreted as each point being a mixture of two materials. Over time, the mixture separates. The simulations are run using py-pde \cite{zwicker-pypre}, with an adaptive time step initialized to $1\times 10^{-4}$. The simulations are run to an end time of $6$. The timestep increases as the simulation runs, for a total of around 300 timesteps.

During training, simulations are run using a relatively coarse $32\times 32$ resolution grid. We use $70\%$ of these simulations for training and the remaining $30\%$ are used as a holdout validation set. The test data set is generated by running groups of 2000 simulations with resolutions of $32 \times 32$, $64 \times 64$, and $128 \times 128$. This gives a total of $6,000$ test problems.

\subsection{Baseline Hyperparameters}

For each model, we perform a grid search over important hyperparameters tuned and we select the model that achieved the lowest validation MSE for testing. Table \ref{tab:ac-hyperparameters} lists the hyperparameters tuned for each baseline model. Each model was trained for $270$ epochs using a cosine annealing learning rate schedule, with $500$ iterations of linear warmup to the base learning rate. The learning rate decays to a final learning rate of $1\times 10^{-8}$. All models are trained with a gradient norm clipping of $0.5$, a batch size of $128$,  the AdamW optimizer with its default weight decay of $0.01$ (except FNO and FFNO, where we tuned a lower setting) \cite{loshchilov2017fixing}. We found that the choice of weight decay had very little effect on the validation metrics.

For the model using Galerkin-style attention, we experiment with two kinds of position encoding. The first, which we call ``concat'' simply concatenates each point's coordinate before applying the lifting operation.  The second, which we call ``add'' uses an MLP to compute a position encoding for each coordinate. This position encoding is added onto the data before each transformer block.

\begin{table}[htbp]
    \centering
    \begin{tabular}{c|c|c}
    Model & Hyperparameter & Value \\
    \hline
        FNO & learning rate & \textbf{0.001}, 0.0005 \\
            & modes & \textbf{16}, 24 \\
            & channels & \textbf{16}, 32 \\
            & weight decay & 0.01, \textbf{0.0001} \\
        \hline
        FFNO & learning rate & \textbf{0.001}, 0.0005 \\
             & modes & 12, \textbf{16}, 24 \\
             & channels & 16, \textbf{32}, 64, 128 \\
        & weight decay & \textbf{0.01}, 0.0001 \\
        & layers & 4, \textbf{5} \\
        \hline
        GT & learning rate & 0.001, \textbf{0.0005} \\
           & embedding dim & 64, 128, \textbf{256}, 512 \\
           & heads & \textbf{4}, 8 \\
           & position enc. & concat, \textbf{add} \\ 
        \hline
        FF & learning rate & 0.001, \textbf{0.0005} \\
           & embedding dim & \textbf{64}, 128, 256, 512 \\
           & heads & \textbf{4}, 8 \\
           & position enc. & Rotary
    \end{tabular}
    \caption{\textbf{Allen-Cahn Hyperparameters in Baselines.} We performed a sweep over the listed settings. The hyperparameters that resulted in the best validation accuracy are marked in bold font. The most important parameters are the modes, channel/embedding dim, and the learning rate.}
    \label{tab:ac-hyperparameters}
\end{table}

Table \ref{tab:ac-in-dist-test-metrics} reports the MAE and MSE for the in-distribution test set.

\subsection{ViT-NO and Swin-NO Hyperparameters}

When tuning our own models, we focus on settings that are specific to our proposed modules. The generic hyperparameter settings we use are identical to the baseline models.  We use AdamW's default learning rate of $0.001$ and
weight decay of $0.01$. We always use a subdomain size of $1/8\times 1/8$, so that our domain is broken into 64 subdomains. Each subdomain has a $4\times 4$ resolution at training and $16\times 16$ resolution on the full $128\times 128$ discretization. We prioritize tuning hyperparameters specific to the subdomain operators, as we either propose the method ourselves or there is no literature describing strong default options. These are listed in Table \ref{tab:ac-experiment-hyperparameters}. 

For the mixture operator, we tune the number of matrices and the embedding dimension. We do not tune the size of the coefficient network $C$ and just set it to an extremely small two layer network with sizes $(4, 8), (8, n)$. The input is two spatial coordinates, so it projects a four-dimensional vector to an $n$-dimensional vector. The hyperparameters we considered are listed in Table \ref{tab:ac-experiment-hyperparameters}

We also consider two additional variants of the ViT operator: the neighborhood operator and a Swin operator. The neighborhood operator reuses the hyperparameters of the best performing ViT model. 

\begin{table}[h]
    \centering
    \begin{tabular}{c|c|c}
        MO & size $n$ & 32, \textbf{64}, 128 \\
        & embedding dim & 32, \textbf{64}, 128, 256\\
        \hline
        Separable MO & size $n$ & 32, 64, 128, \textbf{256} \\
        & embedding dim & 32, 64, \textbf{128}  \\
    \end{tabular}
    \caption{\textbf{Allen Cahn Additional Hyperparameters.} We tune the number of matrices $n$ in the mixture operator (MO) summation and the embedding dimension}
    \label{tab:ac-experiment-hyperparameters}
\end{table}

The Swin-NO-MO uses a window size of $4\times 4$ subdomains and performs shifting across 2 subdomains.

\subsection{Results and Discussion}

\begin{table}[h]
    \centering
    \begin{tabular}{l|c|c|c|c}
\toprule
    \textbf{Model} & \textbf{Metric} & $\mathbf{32\times 32}$ & $\mathbf{64 \times 64}$ & $\mathbf{128\times 128}$  \\
\hline
   FNO & MAE &  0.0154 & 0.0157 & 0.0158 \\
    & MSE & $8.27\times 10^{-4}$ & $8.57\times 10^{-4}$ & $8.90\times 10^{-4}$ \\
\hline
   FFNO & MAE & $4.35\times 10^{-3}$ & $5.01\times 10^{-3}$ & $5.11\times 10^{-3}$  \\
    & MSE & $6.38 \times 10^{-5}$ & $1.06 \times 10^{-4}$ & $1.22 \times 10^{-4}$\\
\hline
   GT & MAE & 0.012 & 0.0122 & 0.0126 \\
   & MSE & $8.3 \times 10^{-4}$ & $8.5 \times 10^{-4}$ & $9.8 \times 10^{-4}$ \\
\hline
   FF & MAE & $5.8 \times 10^{-3}$ & $6.4 \times 10^{-3}$ & $6.6 \times 10^{-3}$ \\
   & MSE & $1.23 \times 10^{-4}$ & $1.59 \times 10^{-4}$ & $1.68 \times 10^{-4}$\\
\hline
   ViT-NO & MAE & $3.58\times10^{-3}$ & $6.98\times10^{-3}$ & $8.12\times 10^{-3}$ \\
   & MSE & $4.42\times 10^{-5}$ & $1.43 \times 10^{-4}$ & $1.95 \times 10^{-4}$  \\
\hline 
   Nbr-NO & MAE & $6.23 \times 10^{-3}$ & $1.6 \times 10^{-2}$ & $1.96 \times 10^{-2}$ \\ 
   & MSE & $1.36 \times 10^{-4}$ & $7.59 \times 10^{-4}$ & $1.14 \times 10^{-3}$ \\
\hline
   Swin-NO & MAE & $4.33\times 10^{-3}$ & $8.01\times 10^{-3}$ & $9.15\times 10^{-3}$\\
   & MSE & $8.82\times 10^{-5}$ & $1.85\times 10^{-4}$ & $1.79\times 10^{-4}$\\
\bottomrule
\end{tabular}
    \caption{\textbf{Allen-Cahn In-Distribution Test Errors.} These models reported achieved the best validation accuracy in our hyperparameter sweep. The training uses $32\times 32$ grid. We test on unseen problems with $32\times 32$, $64\times 64$, and $128\times 128$ resolutions. This test essentially asserts that the models can interpolate on resolutions not seen in training. Best performing ViT-NO, Nbr-NO, and Swin-NO models are with Separable MO, Separable MO, and MO subdomain operators respectively.}
    \label{tab:ac-in-dist-test-metrics}
\end{table}

Table \ref{tab:ac-in-dist-test-metrics} reports the results at training resolution (32$\times$32) and unseen higher resolution (128$\times$128).

\textbf{FNO and FFNO.} We discuss these models together since their results are so similar. We believe that FFNO performed better than FNO because we are able to use a larger hidden dimension, while maintaining a similar parameter count. We found that the choice of weight decay used with AdamW had very little effect on the validation error. 

\textbf{Pure Galerkin Transformer.} We considered a model purely consisting of encoders using Galerkin-style self-attention. We did observe some pathologies when attempting to train larger versions of the Galerkin transformer. While performing hyperparameter tuning, almost every model that used an embedding dimension of 512 experienced extremely large spikes in the validation error that the model was not able to recover from. In the few cases where it did not spike, the accuracy was not competitive.

\textbf{FactFormer.} We notice that FactFormer is competitive with our transformer models. Interestingly, it performs slightly worse than our best transformers, but acheives a very stable interpolation error as it increases to higher resolution. The difference between the MAE at the $32\times 32$ and $128\times 128$ resolutions is only $0.8\times 10^{-3}$.

\textbf{Neighborhood and Swin-NO.} We observe that using neighborhood attention or sliding window attention does not improve the results, even with similar parameter counts to the VIT-NO-MO. On the one hand, this is surprising because the Allen-Cahn equation over this time range should primarily depend on local features. On the other hand, it seems likely that the domain of influence is sufficiently large that restricting the size of the attention operation loses important information.
\section{Navier-Stokes Experiments} \label{app:ns-exp}

\subsection{Hyperparameters}

We follow similar training strategies for Navier-Stokes and Allen-Cahn and reuse many of the same hyperparameters. All models are trained using the AdamW optimizer. We use a linear learning rate warmup of 500 iterations increasing to a learning rate of $0.001$. After warmup, the learning rate decays linearly to to $1e-8$. The models are trained for a total of 100,000 steps. For Navier-Stokes experiment on the Swin-NO, we found that decomposing the domain into a $32\times32$ grid of subdomains improved the results. The subdomains are grouped into blocks of $16\times 16$ subdomains to form windows. In each layer, we perform subdomain shifting across 4 subdomains. For the ViT-NO, we use the same $8\times 8$ grid of subdomains as we did not see a similar improvement when increasing the number of subdomains.

\subsection{Results and Discussion}

\begin{table}[h]
    \centering
    \begin{tabular}{l|c|c|c}
\toprule
    \textbf{Model}& \textbf{Metric} & Error & Rollout Error \\
\hline
   FNO & MAE & $1.27\times10^{-1}$ & $1.72\times 10^{-1}$\\
   & MSE & $3.17\times10^{-2}$ & $5.33\times10^{-2}$\\
\hline
   FFNO & MAE & $5.05\times 10^{-2}$ & $7.39\times10^{-2}$\\
   & MSE & $6.19\times10^{-3}$ & $1.26\times10^{-2}$\\
\hline
   GT & MAE & $1.80\times10^{-1}$ & $2.07\times10^{-1}$\\
   & MSE & $6.13\times10^{-2}$ & $8.11\times10^{-2}$\\
\hline
   FF & MAE & $1.37\times 10^{-1}$ & $1.95\times10^{-1}$ \\
   & MSE & $3.79\times 10^{-2}$ & $7.15\times 10^{-2}$\\
\hline
   ViT-NO & MAE & $1.29\times 10^{-1}$ & $2.63\times 10^{-1}$\\
   & MSE & $4.39\times 10^{-2}$ & $1.47\times 10^{-1}$\\
\hline 
   Nbr-NO & MAE & $1.61\times 10^{-1}$ & $2.81\times 10^{-1}$\\ 
   & MSE & $5.74 \times 10^{-2}$ & $1.46\times 10^{-1}$\\
\hline
   Swin-NO & MAE & $7.07\times10^{-2}$ & $1.33\times10^{-1}$ \\
   & MSE & $1.44\times10^{-2}$  & $4.56\times10^{-2}$\\
\bottomrule
\end{tabular}
\caption{\textbf{Navier-Stokes Rollout Errors.} Best performing ViT-NO, Nbr-NO, and Swin-NO models are with Interpolating, Separable MO, and MO subdomain operators respectively.}
\label{tab:ns-rollout}
\end{table}

\begin{figure}[h]
    \centering
    \includegraphics[width=0.6\linewidth]{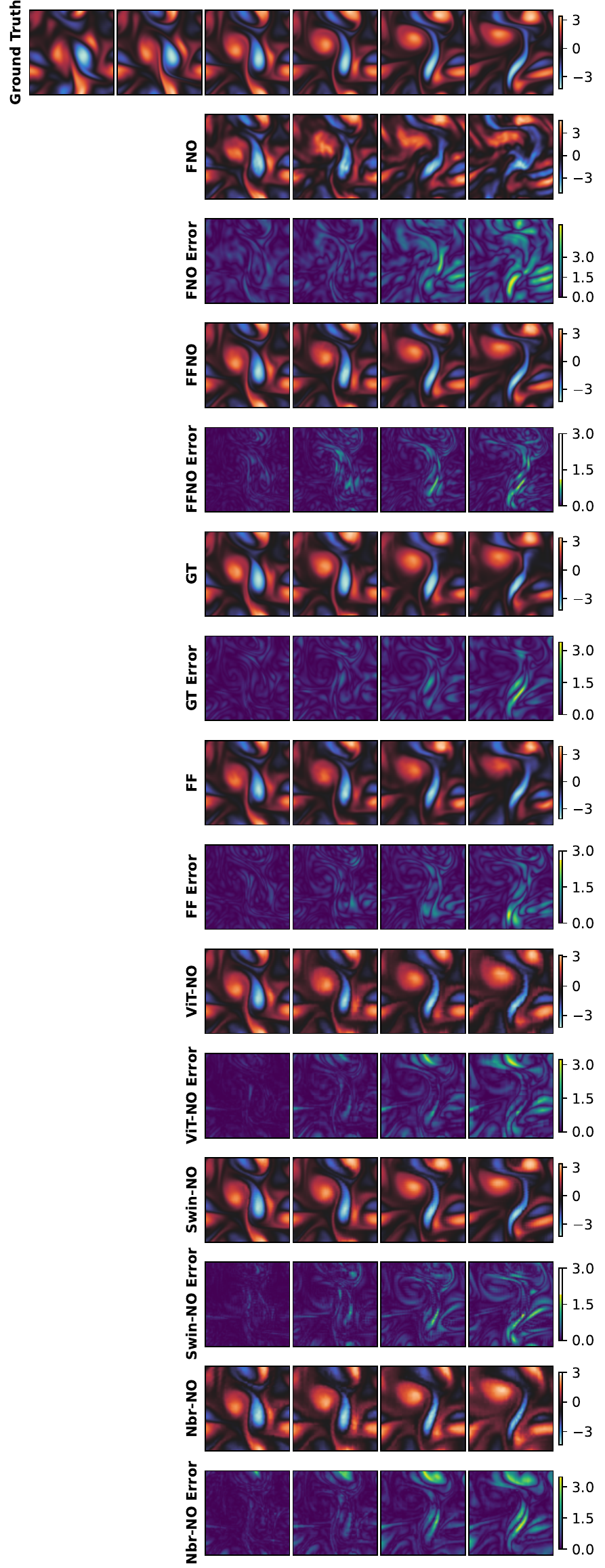}
    \caption{\textbf{Visualization of Navier-Stokes Rollout.}}
    \label{fig:ns-rollout-all-baselines}
\end{figure}

Table \ref{tab:ns-rollout} reports both two-step prediction and rollout errors for various operator learning models on the 2D decaying turbulence problem and Figure \ref{fig:ns-rollout-all-baselines} visualizes the solution.
The Swin-NO using the full-rank mixture operator outperforms ViT-NO and Nbr-NO by a large margin. This validates the architectural insight that shifted window attention is better suited for spatially localized dynamics such as vorticity in turbulence, providing better inductive bias for capturing local flow patterns.

Among baseline models, FFNO performs best, outperforming the original FNO, Galerkin Transformer, and Factformer. The superior performance of FFNO is likely due to its axis-wise decoupling of Fourier modes, which aligns well with the dominant directional structures in 2D turbulence.
FNO fails to capture temporal dynamics robustly here, likely due to spectral truncation that removes high-frequencies. This is consistent with observations that spectral methods tend to oversmooth solutions and suffer from aliasing artifacts in chaotic regimes.

\section{Subdomain Operator Ablations} \label{app:ablations}

We ablate the different subdomain operations in Section \ref{subsec:model:subdomop}.

\subsection{Allen-Cahn} 
\begin{figure}[h]
    \centering
    \includegraphics[width=0.55\linewidth]{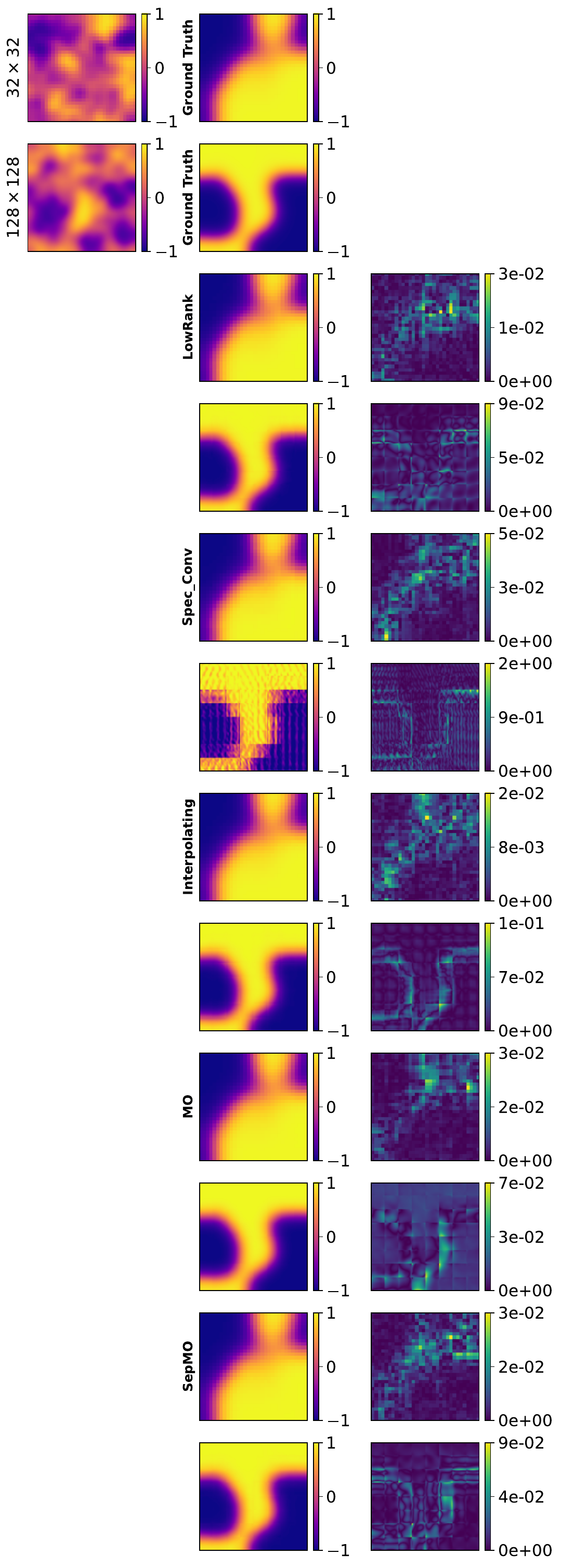}
    \caption{Allen-Cahn ViT Operator}
    \label{fig:ac-vit-subdomain-operator}
\end{figure}

\begin{figure}[h]
    \centering
    \includegraphics[width=0.55\linewidth]{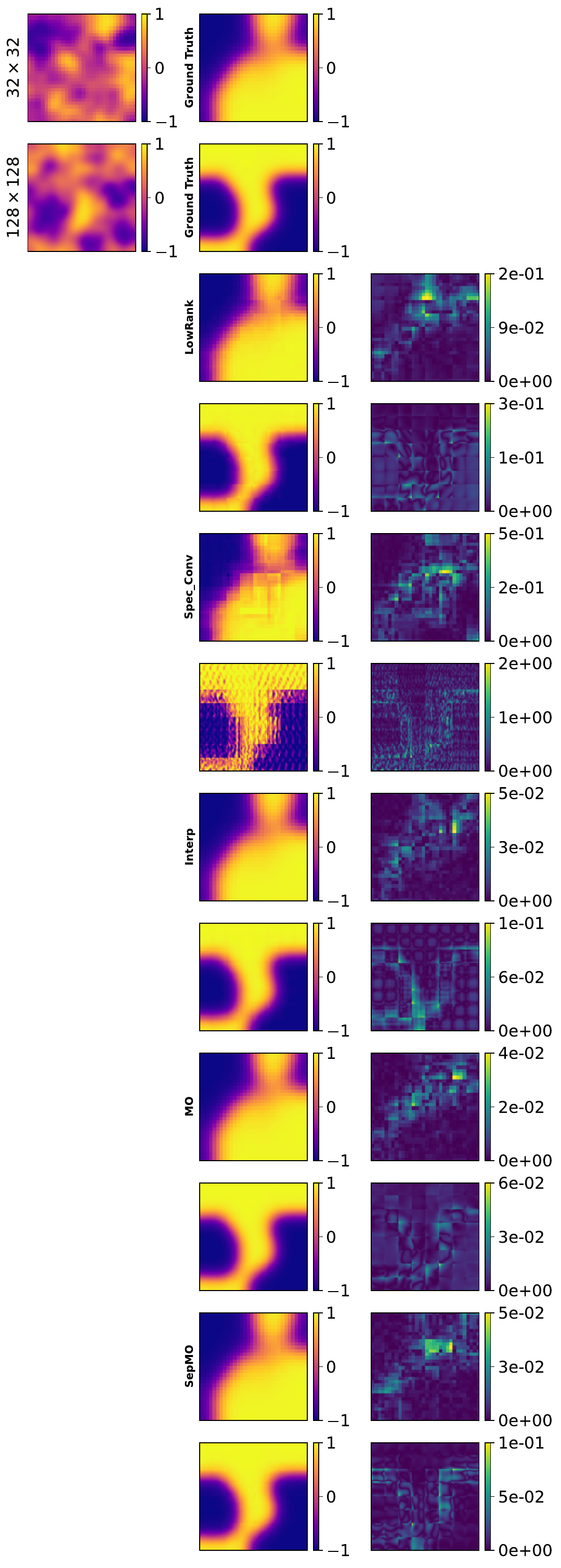}
    \caption{Allen-Cahn Swin Operator}
    \label{fig:ac-swin-subdomain-operator}
\end{figure}

We evaluate different subdomain operators for the ViT-NO and the Swin-NO, with the suffix denoting the choice of subdomain operator in Table \ref{tab:ac-subdomain-ops} and Figures \ref{fig:ac-vit-subdomain-operator} and \ref{fig:ac-swin-subdomain-operator}.

At 32$\times$32, the Mixture Operator (MO) and the Separable Mixture Operator (SepMO) consistently yield the lowest errors for both ViT and Swin backbones. These results highlight that combining multiple parameter matrices using learned coefficients provides a highly expressive yet stable operator for subdomain transformations. SepMO achieves comparable or even lower errors than MO despite $5\times$  fewer parameters, suggesting a strong accuracy-efficiency tradeoff.

In contrast, operators like Spectral Convolution (SpecConv) exhibit worse errors at this resolution. The spectral operator's downsampling and subsequent interpolation likely discard high-frequency interface details important to model Allen-Cahn dynamics, even with skip connections. 
The poor performance of Spectral Convolution across both architectures underscores that downsampling-based spectral filters are less suited for small subdomain resolutions, where truncation of high frequencies may eliminate sharp features (e.g., phase boundaries). Attention-based subdomain operators also underperform, likely due to their softmax smoothing behavior and limited expressivity when constrained to small local neighborhoods.
Interpolating integral operator maintains acceptable errors at 64$\times$64 but degrade substantially at 128$\times$128. This drop can be attributed to the projection-like behavior of such operators, which limits their expressivity on unseen fine-scale features.

\begin{table}[h]
    \centering
    \begin{tabular}{l|c|c|c|c|c}
\toprule
    \textbf{Model} & \textbf{Parameters} & \textbf{Metric} & $\mathbf{32\times 32}$ & $\mathbf{64 \times 64}$ & $\mathbf{128\times 128}$  \\
\hline
   ViT-NO-Vanilla & 6,947,201 & MAE & $6.87\times10^{-3}$ & $4.22\times10^{-2}$ & $5.20\times 10^{-2}$ \\
   & & MSE & $2.25\times 10^{-4}$ & $1.99 \times 10^{-3}$ & $2.91 \times 10^{-3}$  \\
\hline
   ViT-NO-LowRank & 3,078,641 & MAE & $7.16\times10^{-3}$ & $1.13\times10^{-2}$ & $1.29\times10^{-2}$ \\
   & & MSE & $2.47\times10^{-4}$ & $4.88\times10^{-4}$ & $6.13\times10^{-4}$ \\
\hline
   ViT-NO-Attention & 248,403 & MAE & $1.51\times10^{-2}$ & $1.83\times10^{-2}$ & $1.94\times10^{-2}$ \\
   & & MSE & $1.38\times10^{-3}$ & $1.90\times10^{-3}$ & $2.21\times10^{-3}$ \\
\hline
   ViT-NO-SpecConv & 8,542,850 & MAE & $1.85\times10^{-2}$ & $2.10\times10^{-1}$ & $2.14\times10^{-1}$ \\
   & & MSE & $3.36\times10^{-3}$ & $1.03\times10^{-1}$ & $1.17\times10^{-1}$ \\
\hline
   ViT-NO-Interpolating & 3,458,178 & MAE & $3.68 \times 10^{-3}$ & $1.78 \times 10^{-2}$ & $2.03 \times 10^{-2}$\\
   & & MSE & $4.6\times 10^{-5}$ & $9.8 \times 10^{-4}$ & $1.25 \times 10^{-3}$ \\
\hline 
   ViT-NO-MO & 6,954,850 & MAE & $3.66\times 10^{-3}$ & $1.11 \times 10^{-2}$ & $1.30 \times 10^{-2}$ \\
   & & MSE & $4.03 \times 10^{-5}$ & $2.2 \times 10^{-4}$ & $3.02 \times 10^{-4}$ \\
\hline
   ViT-NO-SepMO & 1,331,363 & MAE & $3.58\times10^{-3}$ & $6.98\times10^{-3}$ & $8.12\times 10^{-3}$ \\
   & & MSE & $4.42\times 10^{-5}$ & $1.43 \times 10^{-4}$ & $1.95 \times 10^{-4}$  \\
\hline
\hline
   Swin-NO-Vanilla & 6,945,921 & MAE & $5.74\times 10^{-3}$ & $3.74\times 10^{-2}$ & $4.59\times 10^{-2}$\\
   & & MSE & $1.52\times 10^{-4}$ & $1.71\times 10^{-3}$ & $2.49\times 10^{-3}$\\
\hline
   Swin-NO-LowRank & 3,077,361 & MAE & $2.41\times 10^{-2}$ & $2.50\times 10^{-2}$ & $2.64\times 10^{-2}$\\
   & & MSE & $2.25\times 10^{-3}$ & $2.51\times 10^{-3}$ & $2.76\times 10^{-3}$\\
\hline 
   Swin-NO-Attention & 247,123 & MAE & $1.65\times 10^{-2}$ & $1.81\times 10^{-2}$ & $1.88\times 10^{-2}$\\
   & & MSE & $1.52\times 10^{-3}$ & $1.73\times 10^{-3}$ & $1.84\times 10^{-3}$\\
\hline 
   Swin-NO-SpecConv & 8,541,570 & MAE & $1.32\times 10^{-2}$ & $1.94\times 10^{-1}$ & $2.00\times 10^{-1}$\\
   & & MSE & $1.28\times 10^{-3}$ & $8.76\times 10^{-2}$ & $9.49\times 10^{-2}$\\
\hline 
   Swin-NO-MO & 6,953,570 & MAE & $4.33\times 10^{-3}$ & $8.01\times 10^{-3}$ & $9.15\times 10^{-3}$\\
   & & MSE & $8.82\times 10^{-5}$ & $1.85\times 10^{-4}$ & $1.79\times 10^{-4}$\\
\hline
   Swin-NO-SepMO & 1,109,410  & MAE & $9.40\times 10^{-3}$ & $1.17\times 10^{-2}$ & $1.28\times 10^{-2}$\\
   & & MSE & $5.05\times 10^{-4}$ & $7.33\times 10^{-4}$ & $5.83\times 10^{-4}$\\
\bottomrule
\end{tabular}
    \caption{\textbf{Comparison of Subdomain Operators for Allen-Cahn.}}
    \label{tab:ac-subdomain-ops}
\end{table}

\subsection{Navier-Stokes.} 

\begin{figure}[h]
    \centering
    \includegraphics[width=0.7\linewidth]{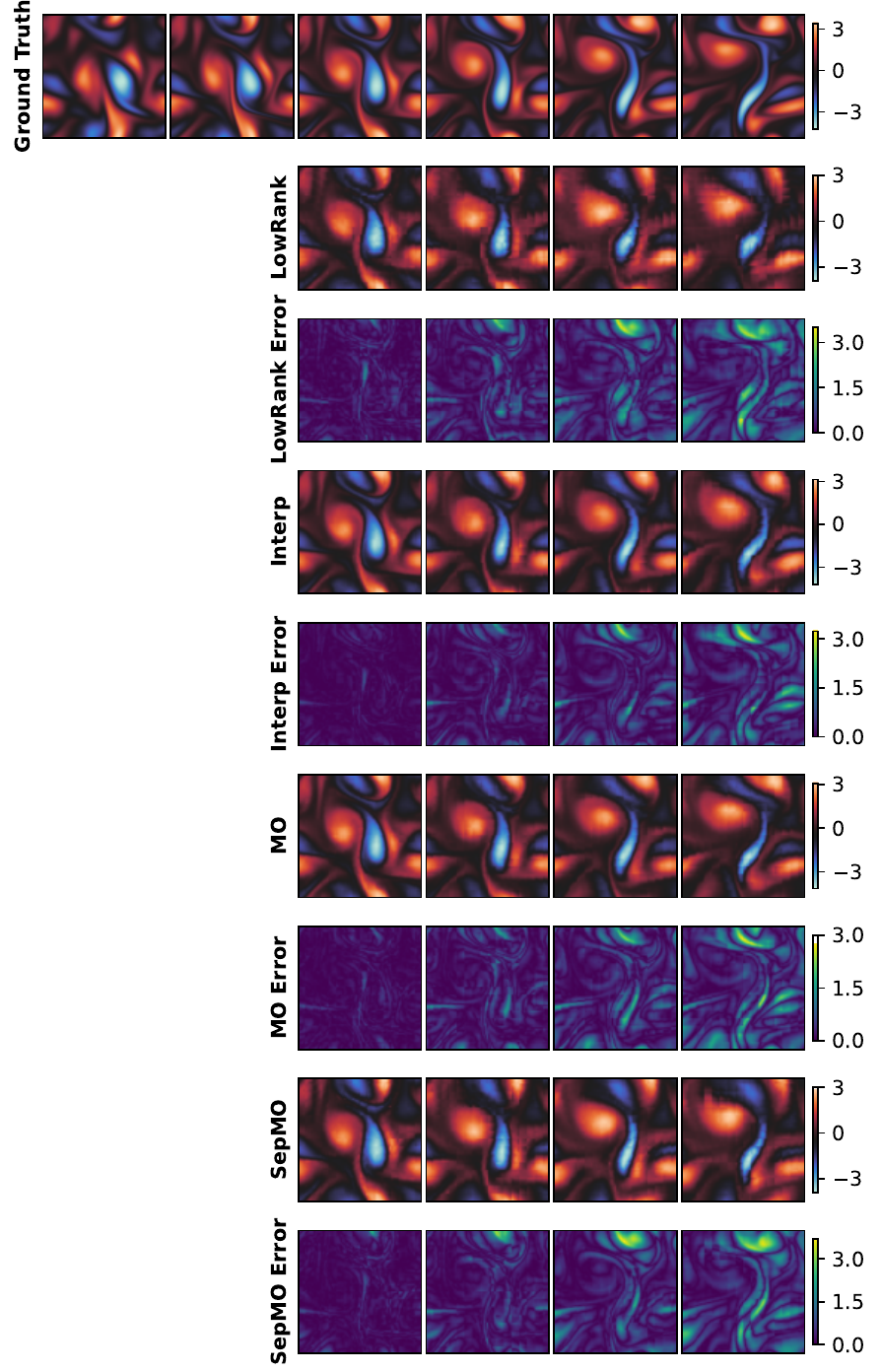}
    \caption{\textbf{Visualization of different subdomain operators for ViT-NO Navier-Stokes Rollout. These plots show the absolute error.}}
    \label{fig:ns-rollout-vit-subdomain-ops}
\end{figure}

\begin{figure}[h]
    \centering
    \includegraphics[width=0.7\linewidth]{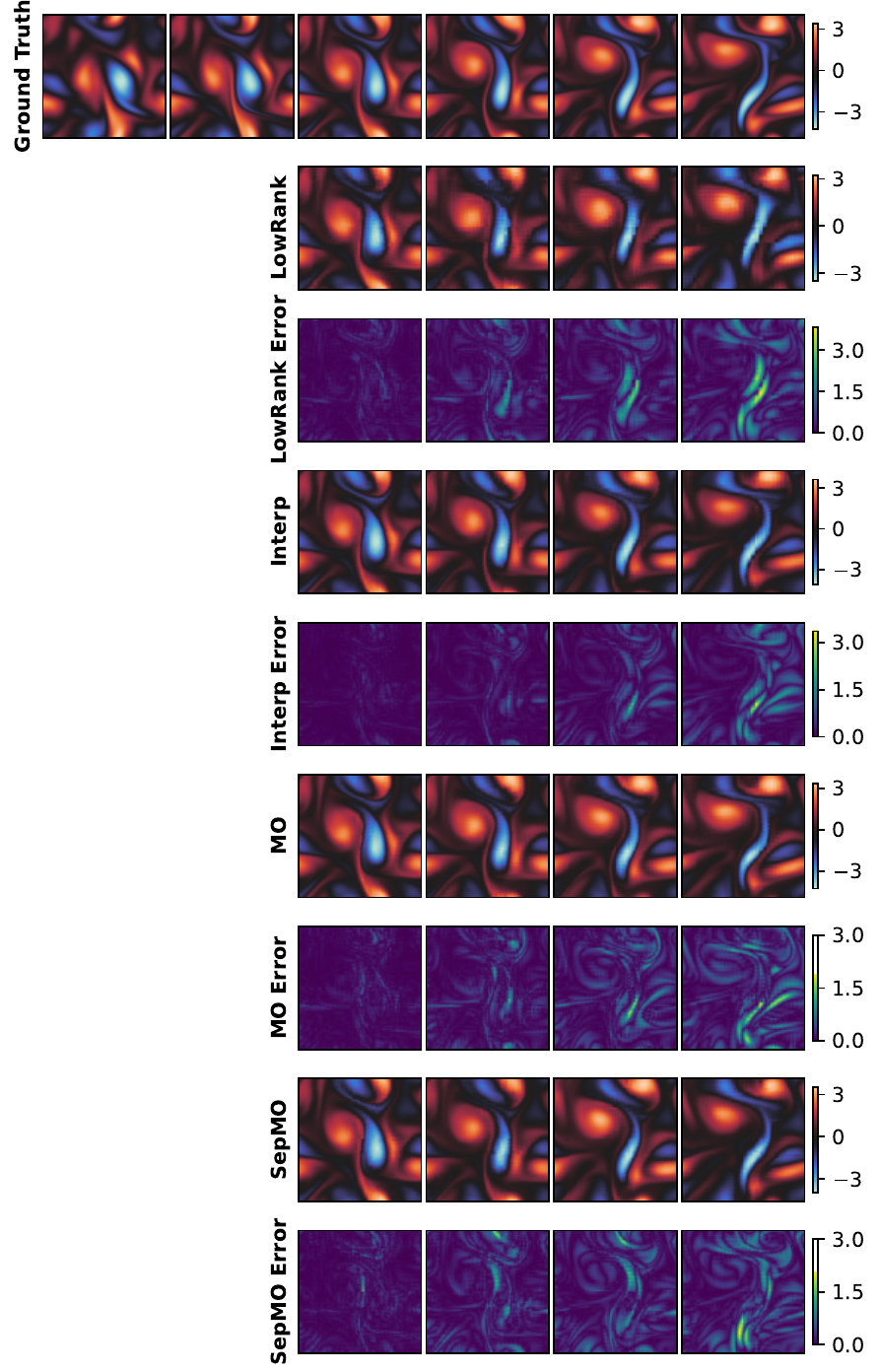}
    \caption{\textbf{Visualization of different subdomain operators for Swin-NO Navier-Stokes Rollout.  These plots show the absolute error.} 
    }
    \label{fig:ns-rollout-swin-subdomain-ops}
\end{figure}

The Navier-Stokes equation models chaotic, multi-scale dynamics, making it a challenging test for subdomain operator design. Unlike Allen-Cahn, errors compound during rollouts, emphasizing the importance of stability. Table \ref{tab:ns-subdomain-ops} compares the performance of various subdomain operators across two metrics: two-step prediction error and multi-step rollout error. Figures \ref{fig:ns-rollout-vit-subdomain-ops} and \ref{fig:ns-rollout-swin-subdomain-ops} visualize the rollout of the different subdomain operators with ViT and Swin backbones. 

The Mixture Operator paired with the Swin backbone outperforms all other configurations, achieving the lowest MAE and MSE for single-step prediction and maintaining strong stability during rollouts. This confirms earlier observations from Allen-Cahn: combining learned kernel matrices via a coefficient network leads to high expressivity without sacrificing generalization. The Swin architecture's hierarchical attention and shift operations further improve feature aggregation across time-evolving turbulence.

\begin{table}[h]
    \centering
    \begin{tabular}{l|c|c|c|c}
\toprule
    \textbf{Model} & \textbf{Parameters} & \textbf{Metric} & Error & Rollout Error \\
\hline
   ViT-NO-LowRank & 3,091,970 & MAE & $2.14\times10^{-1}$ & $3.65\times 10^{-1}$\\
   & & MSE & $8.77\times10^{-2}$ & $2.08\times 10^{-1}$\\
\hline
   ViT-NO-Attention & 260,756 & MAE & $1.64\times10^{-1}$ & $3.09\times10^{-1}$ \\
   & & MSE & $6.23\times10^{-2}$ & $1.74\times10^{-1}$\\
\hline
   ViT-NO-Interpolating & 3,483,396 & MAE & $1.29\times 10^{-1}$ & $2.63\times 10^{-1}$\\
   & & MSE & $4.39\times 10^{-2}$ & $1.47\times 10^{-1}$\\
\hline 
   ViT-NO-MO & 6,971,300 & MAE & $1.44\times 10^{-1}$  & $2.79\times 10^{-1}$\\
   & & MSE & $5.09\times 10^{-2}$ & $1.52\times 10^{-1}$\\
\hline
   ViT-NO-SepMO & 1,136,676 & MAE & $1.85\times10^{-1}$ & $3.27\times 10^{-1}$\\
   & & MSE & $7.19\times10^{-2}$ & $1.84\times 10^{-1}$\\
\hline 
\hline
   Swin-NO-LowRank & 3,093,762  & MAE & $1.55\times10^{-1}$ & $2.50\times10^{-1}$ \\
   & & MSE & $5.53\times10^{-2}$  & $1.20\times10^{-1}$\\
\hline
   Swin-NO-Attention & 247,188 & MAE & $1.56\times10^{-1}$ & $2.78\times10^{-1}$\\
   & & MSE & $5.54\times10^{-2}$ & $1.43\times10^{-1}$\\
\hline
   Swin-NO-SpecConv & 2,568,580 & MAE & $1.59\times10^{-1}$ & $2.60\times10^{-1}$\\
   & & MSE & $5.25\times10^{-1}$ & $1.26\times10^{-1}$\\
\hline 
   Swin-NO-Interpolating & 1,638,020  & MAE & $7.71\times10^{-2}$ & $1.46\times10^{-1}$ \\
   & & MSE &  $1.70\times10^{-2}$ & $5.43\times10^{-2}$\\
\hline 
   Swin-NO-MO & 6,973,092 & MAE & $7.07\times10^{-2}$ & $1.33\times10^{-1}$ \\
   & & MSE & $1.44\times10^{-2}$  & $4.56\times10^{-2}$\\
\hline
   Swin-NO-SepMO & 1,109,540 & MAE & $1.23\times 10^{-1}$ & $1.79\times 10^{-1}$\\
   & & MSE & $3.66 \times 10^{-2}$ & $6.94 \times 10^{-2}$\\
\bottomrule
\end{tabular}
\caption{\textbf{Comparison of Subdomain Operators for Navier-Stokes.}}
\label{tab:ns-subdomain-ops}
\end{table}

\end{document}